\documentclass[preprint,12pt]{elsarticle}

\usepackage{amsmath,amssymb,amsfonts}
\usepackage{booktabs}
\usepackage{graphicx}
\usepackage{array}
\usepackage{multirow}
\usepackage{placeins}
\usepackage{hyperref}
\usepackage{tikz}
\usetikzlibrary{arrows.meta,calc,fit,positioning,backgrounds}

\journal{Neurocomputing}

\newcommand{\R}{\mathbb{R}}
\newcommand{\Dphi}{D_{\phi}}
\newcommand{\PB}{Role-Aware Bregman}
\newcommand{\ICNN}{ICNN-Bregman}

\begin{document}

\begin{frontmatter}

\title{Role-Aware Neural Convex Divergence Heads for Asymmetric Representation Learning}

\author[inst1,inst2]{He Huang\fnref{fn1}\corref{cor1}}
\ead{huanghe@ctbu.edu.cn}
\author[inst3]{Lu Shen\fnref{fn1}}
\author[inst4]{Yunfeng Huang}
\author[inst1,inst2]{Li Qi}
\fntext[fn1]{He Huang and Lu Shen contributed equally to this work.}
\cortext[cor1]{Corresponding author.}
\address[inst1]{School of Mathematics and Statistics, Chongqing Technology and Business University, Chongqing, 400067, China}
\address[inst2]{Chongqing Key Laboratory of Statistical Intelligent Computing and Monitoring, Chongqing Technology and Business University, Chongqing, 400067, China}
\address[inst3]{School of Food Science, Chongqing Technology and Business University, Chongqing, 400067, China}
\address[inst4]{Faculty of Electrical Engineering and Information Technology, TU Dortmund University, Dortmund, Germany}

\begin{abstract}
Many representation learning problems involve directed relations, such as lexical entailment, sentence entailment, ontology hierarchy, and citation links. Standard Euclidean, cosine, and Mahalanobis heads are symmetric, while generic neural scorers can model directionality but provide limited geometric structure. This paper proposes a role-aware neural convex divergence head for asymmetric representation learning. The head applies source- and target-role projections before evaluating an input-convex neural Bregman divergence, yielding a nonnegative structured score in the role-projected space. We characterize its projected-space identity, source-role convexity, directional-gap decomposition, and Hessian-based local curvature. Experiments on lexical, sentence, ontology, and directed graph benchmarks compare symmetric distances, unstructured asymmetric scorers, order/hyperbolic baselines, plain ICNN-Bregman heads, and the proposed role-aware variant. Across ten random seeds on the main semantic and ontology benchmarks, role-aware projections consistently improve directional accuracy over plain ICNN-Bregman heads while preserving zero observed negative divergence rate. The results also identify a boundary case: on large fixed-feature citation prediction, specialized symmetric or hyperbolic baselines remain stronger in ranking accuracy. Overall, the proposed head is best understood as a structured and interpretable plug-in distance module for tasks where directional relations matter.
\end{abstract}


\end{frontmatter}

\section{Introduction}

Distance functions are a central component of representation learning. They are used to compare examples in metric learning, retrieve nearest neighbors, score candidate links, classify examples by prototypes, and regularize embedding spaces. A large fraction of practical systems still rely on symmetric distances such as Euclidean distance, cosine distance, or Mahalanobis distance. These choices are effective when the relation of interest is similarity. However, many relations studied in machine learning are not symmetric. Hypernymy is directed from a specific concept to a more general concept; entailment is directed from a premise to a hypothesis; ontology edges are directed from child terms to parent terms; and citation links point from a citing paper to a cited paper.

The importance of asymmetry in similarity judgments is not new: Tversky's feature-based theory showed that similarity can be directional and context-dependent, challenging purely metric views of comparison \citep{tversky1977features}. In representation learning, one response is to abandon distance structure and use a generic neural scorer, such as a multilayer perceptron on concatenated pair features. This provides expressive asymmetry but weakens the geometric meaning of the learned score. Another response is to use structured asymmetric geometry, including order embeddings \citep{vendrov2016order} or hyperbolic embeddings \citep{nickel2017poincare}. These approaches are well matched to hierarchical data, but their assumptions can be restrictive and they are usually implemented as a complete embedding model rather than as a plug-in head for arbitrary encoders.

This paper studies a middle ground. We ask whether an asymmetric head can be both neural and geometrically structured. Our starting point is the Bregman divergence, originally introduced in convex programming \citep{bregman1967relaxation},
\begin{equation}
\Dphi(x,y)=\phi(x)-\phi(y)-\nabla \phi(y)^{\top}(x-y),
\end{equation}
where $\phi$ is a differentiable convex potential. Bregman divergences are generally asymmetric and nonnegative, and they include many classical divergences as special cases \citep{banerjee2005clustering}. Their properties are grounded in classical convex analysis and convex optimization \citep{rockafellar1970convex, boyd2004convex}. Recent work has explored neural parameterizations of Bregman divergences and convex potentials \citep{amos2017input, cilingir2020deep, lu2023neural}. We build on this line of work, but target a specific problem: using neural convex divergences as plug-in distance heads for asymmetric representation learning.

The proposed method, called the role-aware neural convex divergence head, applies two learnable role projections before computing a neural Bregman divergence:
\begin{equation}
D_{\phi,P_s,P_t}(x,y)=D_{\phi}(P_s x, P_t y),
\end{equation}
where $P_s$ and $P_t$ map an input embedding into source and target roles. This construction turns a known asymmetric divergence into a practical, role-aware, encoder-agnostic head. The contribution lies in the combination of a plug-in ICNN potential, role-specific projections, directed training losses, and interpretation tools for asymmetric pairwise learning.

This work makes three main contributions.

First, we propose a role-aware neural convex divergence head for asymmetric representation learning. The head can be attached to fixed embeddings or learned encoders and returns a nonnegative Bregman divergence in a source-target projected space.

Second, we provide a theoretical and diagnostic characterization of the head. We show how classical Bregman properties are retained after role projection, derive a quadratic special-case decomposition of the directional gap, and connect this decomposition to Hessian-based local curvature analysis.

Third, we conduct a systematic empirical study across lexical entailment, sentence entailment, ontology hierarchy, and directed citation tasks. The experiments compare symmetric distances, unstructured asymmetric scorers, order/hyperbolic baselines, plain ICNN-Bregman heads, and the proposed role-aware variant, with ablations, significance tests, and interpretability diagnostics.

\section{Related Work}

\subsection{Metric learning and symmetric distance heads}

Classical metric learning methods learn distances that preserve neighborhood or constraint structure, including large-margin nearest-neighbor learning \citep{weinberger2009distance} and information-theoretic metric learning \citep{davis2007information}. The broader literature is reviewed by \citet{kulis2013metric}, who emphasizes metric learning as the problem of adapting the geometry of a representation space to task-specific comparison constraints. In modern deep learning pipelines, learned encoders are often paired with Euclidean, cosine, or Mahalanobis heads. These heads are stable, efficient, and interpretable for similarity, but they satisfy $d(x,y)=d(y,x)$ and therefore collapse direction-specific information by design.

\subsection{Asymmetric representation learning}

Directed relations have motivated specialized embedding geometries. Order embeddings model partial-order relations through coordinate-wise inequalities and have been used for visual-semantic hierarchy and entailment \citep{vendrov2016order}. Hyperbolic and Poincare embeddings exploit negative curvature to represent trees and hierarchies compactly \citep{nickel2017poincare}. Knowledge-graph models such as TransE and RotatE provide additional examples of structured directional scoring for relational data \citep{bordes2013translating, sun2019rotate}. These models are powerful, but they often impose task-specific geometry or relation-specific scoring assumptions.

\subsection{Input-convex networks and neural Bregman divergences}

Input-convex neural networks (ICNNs) parameterize convex scalar functions by constraining selected weights to be nonnegative \citep{amos2017input}. Convex neural potentials connect to a longer tradition of convex analysis, Bregman distances, and projection methods \citep{bregman1967relaxation, censor1981iterative, rockafellar1970convex}. Deep divergence learning and neural Bregman divergence models show that neural potentials can learn flexible dissimilarities while retaining structure inherited from convexity \citep{cilingir2020deep, siahkamari2020learning, lu2023neural}. Our work differs in emphasis: we treat the divergence as a reusable head for directed representation learning and evaluate whether role-specific projections improve directional behavior without abandoning Bregman structure.

\subsection{Benchmarks for directed semantic and graph relations}

HyperLex evaluates graded lexical entailment and hypernymy \citep{vulic2017hyperlex}. WordNet provides large lexical taxonomies \citep{miller1995wordnet}. SICK and SNLI are standard sentence-pair entailment resources \citep{marelli2014sick, bowman2015snli}. Gene Ontology provides directed biological concept relations \citep{geneontology2019resource}. OGB link-prediction datasets evaluate directed graph prediction under larger-scale fixed-feature regimes \citep{hu2020open}. These resources allow us to test whether the proposed head is useful beyond one narrow dataset.

\section{Method}

\subsection{Problem setting}

Let $x,y\in\R^d$ be two embeddings produced by an encoder or directly provided as fixed input features. A directed pair $(x,y)$ indicates that $x$ should be related to $y$ in a source-to-target direction. The datasets used in this paper do not provide ground-truth divergence values $D(x,y)$. Instead, they provide directed relation labels or graded relation scores, such as child-to-parent ontology edges, premise-to-hypothesis entailment labels, or citing-to-cited paper links. The divergence is therefore a learned scoring function. The goal is to learn a score $s(x,y)$ or distance-like quantity $D(x,y)$ such that positive directed pairs have smaller divergence than corrupted negative pairs, and such that the forward direction can be distinguished from the reverse direction when the task requires it.

The head is designed to be plug-in in two senses. Mathematically, it only consumes embeddings and returns a pairwise divergence, so it can be placed after different encoders. In software, it is implemented as a module with the same interface as ordinary distance heads: given two batches of embeddings, it returns one scalar per pair.

\subsection{Input-convex potential}

We parameterize a differentiable strongly convex potential $\phi_{\theta}:\R^k\rightarrow\R$ using an ICNN \citep{amos2017input}. A typical layer has the form
\begin{equation}
z_{\ell+1}=\sigma(W_{\ell}^{(z)}z_{\ell}+W_{\ell}^{(u)}u+b_{\ell}),
\end{equation}
where $u$ is the input, $\sigma$ is a convex nondecreasing activation, and $W_{\ell}^{(z)}$ is constrained to be element-wise nonnegative. To improve numerical stability and ensure strong convexity, we use a quadratic term:
\begin{equation}
\phi_{\theta}(u)=g_{\theta}(u)+\frac{\lambda}{2}\lVert u\rVert_2^2,\quad \lambda>0,
\end{equation}
where $g_{\theta}$ is the ICNN output.

The induced Bregman divergence is
\begin{equation}
D_{\phi_{\theta}}(u,v)=\phi_{\theta}(u)-\phi_{\theta}(v)-\nabla\phi_{\theta}(v)^{\top}(u-v).
\end{equation}
This quantity is nonnegative when $\phi_{\theta}$ is convex, and it is generally asymmetric.

\subsection{Role-aware projected divergence}

Plain Bregman divergence is asymmetric, but in practice the learned direction may be weak if both arguments are passed through the same representation role. We therefore introduce role projections:
\begin{equation}
u=P_s x,\quad v=P_t y,
\end{equation}
where $P_s$ and $P_t$ are learnable source and target maps. The proposed head is
\begin{equation}
D_{\theta,P_s,P_t}(x,y)
=D_{\phi_{\theta}}(P_sx,P_ty).
\label{eq:roleaware}
\end{equation}
The projections can be linear maps, shallow MLPs, or constrained affine maps. The experiments mainly use linear maps because they give the clearest interpretation: the model learns which directions of the embedding space matter when an item acts as a source and which directions matter when it acts as a target.

\begin{figure}[t]
\centering
\resizebox{\linewidth}{!}{\begin{tikzpicture}[
    font=\sffamily\small,
    >=Latex,
    box/.style={draw=#1, rounded corners=3pt, line width=0.85pt, align=center,
        inner sep=4pt, minimum height=10mm},
    box/.default=black,
    flow/.style={->, line width=0.85pt, draw=black!68},
    soft/.style={-, line width=0.5pt, draw=black!20},
    skip/.style={->, line width=0.7pt, draw=black!55, rounded corners=5pt},
    label/.style={font=\sffamily\small, text=black!72, align=center},
    smalllabel/.style={font=\sffamily\scriptsize, text=black!62, align=center}
]

\definecolor{ink}{RGB}{38,49,63}
\definecolor{line}{RGB}{77,88,102}
\definecolor{panel}{RGB}{247,248,250}
\definecolor{blueFill}{RGB}{226,239,255}
\definecolor{blueLine}{RGB}{92,133,184}
\definecolor{greenFill}{RGB}{232,246,232}
\definecolor{greenLine}{RGB}{94,151,95}
\definecolor{orangeFill}{RGB}{248,211,166}
\definecolor{purpleFill}{RGB}{239,234,255}

\node[label] at (6mm,34mm) {directed pair};
\node[label] at (43mm,34mm) {encoder};
\node[label] at (81mm,34mm) {embeddings};

\node[box=line, fill=white, minimum width=22mm, minimum height=15mm] (x) at (6mm,20mm) {$x$};
\node[box=line, fill=white, minimum width=22mm, minimum height=15mm] (y) at (6mm,0mm) {$y$};

\foreach \i/\yy in {1/22,2/12,3/2} {
    \node[circle, draw=line, fill=white, minimum size=5mm, inner sep=0pt] (e1\i) at (34mm,\yy mm) {};
}
\foreach \i/\yy in {1/26,2/16,3/6,4/-4} {
    \node[circle, draw=line, fill=white, minimum size=5mm, inner sep=0pt] (e2\i) at (47mm,\yy mm) {};
}
\foreach \i/\yy in {1/22,2/12,3/2} {
    \node[circle, draw=line, fill=white, minimum size=5mm, inner sep=0pt] (e3\i) at (60mm,\yy mm) {};
}
\foreach \a in {1,2,3} {
    \foreach \b in {1,2,3,4} {
        \draw[soft] (e1\a) -- (e2\b);
    }
}
\foreach \a in {1,2,3,4} {
    \foreach \b in {1,2,3} {
        \draw[soft] (e2\a) -- (e3\b);
    }
}
\draw[flow] (x.east) -- (e11.west);
\draw[flow] (y.east) -- (e13.west);

\node[box=line, fill=white, minimum width=15mm, minimum height=18mm] (zxTop) at (82mm,20mm) {};
\node[box=line, fill=white, minimum width=15mm, minimum height=18mm] (zyTop) at (82mm,0mm) {};
\foreach \yy in {25,20,15} {\draw[line width=0.8pt, draw=line] (78.5mm,\yy mm) -- (85.5mm,\yy mm);}
\foreach \yy in {5,0,-5} {\draw[line width=0.8pt, draw=line] (78.5mm,\yy mm) -- (85.5mm,\yy mm);}
\draw[flow] (e31.east) -- (zxTop.west);
\draw[flow] (e33.east) -- (zyTop.west);

\node[box=line, fill=panel, minimum width=42mm, minimum height=24mm] (headTop) at (123mm,10mm)
    {\textbf{role-aware}\\\textbf{plug-in head}};
\draw[flow] (zxTop.east) -- ($(headTop.west)+(0,5mm)$);
\draw[flow] (zyTop.east) -- ($(headTop.west)+(0,-5mm)$);

\node[circle, draw=ink, fill=white, line width=1pt, minimum size=14mm] (dTop) at (160mm,10mm) {$D$};
\node[box=line, fill=white, minimum width=25mm, minimum height=13mm] (train) at (191mm,21mm) {train};
\node[box=line, fill=white, minimum width=25mm, minimum height=13mm] (explain) at (191mm,-1mm) {explain};
\draw[flow] (headTop.east) -- (dTop.west);
\draw[flow] (dTop.east) -- (train.west);
\draw[flow] (dTop.east) -- (explain.west);

\node[font=\sffamily\bfseries\large, text=ink] at (111mm,-30mm)
    {zoomed view of the proposed head};
\node[
    draw=ink,
    fill=panel,
    rounded corners=8pt,
    line width=1.05pt,
    minimum width=166mm,
    minimum height=53mm,
    anchor=north west
] (zoom) at (28mm,-34mm) {};

\node[smalllabel] at (36mm,-55mm) {$z_x$};
\node[smalllabel] at (36mm,-78mm) {$z_y$};
\node[box=line, fill=white, minimum width=13mm, minimum height=20mm] (zx) at (50mm,-55mm) {};
\node[box=line, fill=white, minimum width=13mm, minimum height=20mm] (zy) at (50mm,-78mm) {};
\foreach \yy in {-49,-55,-61} {\draw[line width=0.75pt, draw=line] (46.5mm,\yy mm) -- (53.5mm,\yy mm);}
\foreach \yy in {-72,-78,-84} {\draw[line width=0.75pt, draw=line] (46.5mm,\yy mm) -- (53.5mm,\yy mm);}

\node[box=blueLine, fill=blueFill, minimum width=24mm, minimum height=13mm] (ps) at (74mm,-55mm) {$P_s$};
\node[box=greenLine, fill=greenFill, minimum width=24mm, minimum height=13mm] (pt) at (74mm,-78mm) {$P_t$};
\node[smalllabel] at (74mm,-41mm) {source role};
\node[smalllabel] at (74mm,-68mm) {target role};

\node[box=line, fill=white, minimum width=13mm, minimum height=20mm] (uvec) at (97mm,-55mm) {};
\node[box=line, fill=white, minimum width=13mm, minimum height=20mm] (vvec) at (97mm,-78mm) {};
\foreach \yy in {-49,-55,-61} {\draw[line width=0.75pt, draw=line] (93.5mm,\yy mm) -- (100.5mm,\yy mm);}
\foreach \yy in {-72,-78,-84} {\draw[line width=0.75pt, draw=line] (93.5mm,\yy mm) -- (100.5mm,\yy mm);}
\node[smalllabel] at (86mm,-55mm) {$u$};
\node[smalllabel] at (86mm,-78mm) {$v$};

\node[label] at (124mm,-44mm) {input-convex potential};
\foreach \i/\yy in {1/-53,2/-61,3/-69,4/-77} {
    \node[circle, draw=line, fill=blueFill!35, minimum size=5.5mm, inner sep=0pt] (p1\i) at (115mm,\yy mm) {};
}
\foreach \i/\yy in {1/-49,2/-57,3/-65,4/-73,5/-81} {
    \node[circle, draw=line, fill=purpleFill!38, minimum size=5.5mm, inner sep=0pt] (p2\i) at (128mm,\yy mm) {};
}
\foreach \i/\yy in {1/-53,2/-61,3/-69,4/-77} {
    \node[circle, draw=line, fill=orangeFill!55, minimum size=5.5mm, inner sep=0pt] (p3\i) at (141mm,\yy mm) {};
}
\foreach \a in {1,2,3,4} {
    \foreach \b in {1,2,3,4,5} {
        \draw[soft] (p1\a) -- (p2\b);
    }
}
\foreach \a in {1,2,3,4,5} {
    \foreach \b in {1,2,3,4} {
        \draw[soft] (p2\a) -- (p3\b);
    }
}

\node[circle, draw=ink, fill=white, line width=1pt, minimum size=14mm] (dZoom) at (160mm,-65mm) {$D$};
\node[box=line, fill=white, minimum width=24mm, minimum height=12mm] (loss) at (183mm,-55mm) {loss};
\node[box=line, fill=white, minimum width=24mm, minimum height=12mm] (diag) at (183mm,-78mm) {diagnosis};

\draw[flow] (zx.east) -- (ps.west);
\draw[flow] (zy.east) -- (pt.west);
\draw[flow] (ps.east) -- (uvec.west);
\draw[flow] (pt.east) -- (vvec.west);
\draw[flow] (uvec.east) -- (p11.west);
\draw[flow] (vvec.east) -- (p14.west);
\draw[flow] (p32.east) -- (dZoom.west);
\draw[flow] (dZoom.east) -- (loss.west);
\draw[flow] (dZoom.east) -- (diag.west);

\draw[skip] ($(headTop.south)+(0,-1mm)$) -- ++(0,-8mm) -| (zoom.north);
\node[smalllabel, text=black!58] at (105mm,-96mm)
    {The same head can be attached after fixed embeddings or learned encoders.};

\end{tikzpicture}}
\caption{Neural architecture view of the proposed role-aware plug-in divergence head. A directed pair is encoded into two embeddings, the proposed head applies source- and target-role projections, and an ICNN-induced Bregman divergence returns an ordered-pair score. The score is used for pairwise ranking losses and for geometric diagnostics such as directional gaps and local curvature.}
\label{fig:architecture}
\end{figure}
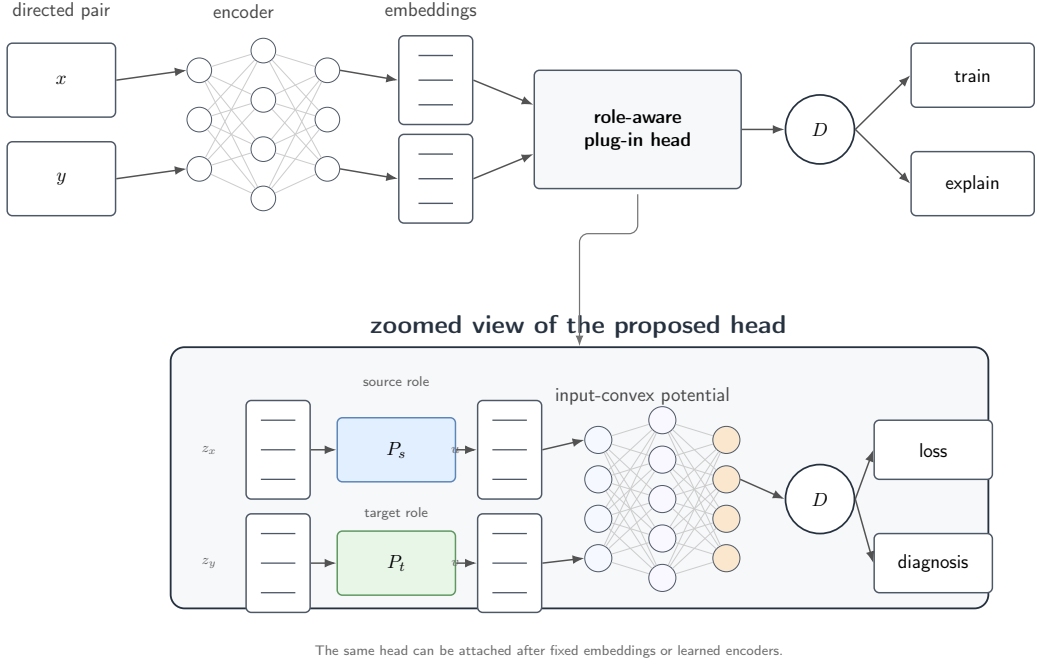

Figure~\ref{fig:architecture} summarizes the system-level role of the proposed head. A pair of items is encoded into two embeddings, the proposed head scores the ordered pair, and the resulting divergence can be used for ranking, link prediction, or interpretation. In the experiments, this scoring module is trained through triplet-style pairwise ranking, described below. Although the reported experiments use fixed embeddings to isolate the head, the module itself has the same input-output interface as ordinary distance heads and can be placed after a learned encoder.

\subsection{Training objectives}

For directed link prediction and pairwise relation learning, the experiments use a triplet-style pairwise ranking objective, following the large-margin metric-learning view that supervision can be given as relative comparison constraints rather than absolute distances \citep{weinberger2009distance, kulis2013metric}. Each observed directed relation $(x,y^+)$ is treated as a positive source-target pair. During training, one corrupted target $y^-$ is sampled for each positive pair from a candidate pool that excludes annotated positive targets of the same source. This negative-sampling protocol is also common in embedding-based link prediction, where models are trained to score observed edges above corrupted edges \citep{bordes2013translating}. We then minimize a margin objective:
\begin{equation}
\mathcal{L}_{rank}
=\max\{0,m + D(x,y^+) - D(x,y^-)\},
\end{equation}
where $(x,y^+)$ is a positive directed pair and $y^-$ is a sampled corrupted target. Thus a minibatch contains triples $(x,y^+,y^-)$, but the learned score remains an ordered-pair divergence $D(x,y)$. The loss does not regress to a known numerical divergence; it only enforces that the learned divergence ranks the annotated relation above sampled non-relations. Across epochs, the same positive pair can be compared with different corrupted targets. For direction supervision, we add a forward-reverse margin:
\begin{equation}
\mathcal{L}_{dir}
=\max\{0,m_d + D(x,y)-D(y,x)\}.
\end{equation}
The full objective is
\begin{equation}
\mathcal{L}=\mathcal{L}_{rank}+\alpha\mathcal{L}_{dir},
\end{equation}
where $\alpha$ controls the ranking-direction trade-off.

Thus, the main training signal is relational: positive pairs should be closer than corrupted pairs, and annotated forward directions should be preferred over their reversals. In the reported experiments, the input representations are held fixed and only the comparison head is optimized. This design isolates the behavior of the proposed head and allows the same training protocol to be applied to heterogeneous datasets whose labels are edges, entailment judgments, or graded relation strengths rather than explicit metric distances.

\section{Theoretical Analysis}

\subsection{Nonnegativity and identity in projected space}

\textbf{Proposition 1.}
If $\phi$ is convex and differentiable, the role-aware head in Eq.~\eqref{eq:roleaware} is nonnegative. If $\phi$ is strictly convex, its zero set is characterized by equality in the role-projected space.

\textit{Proof.}
The first-order supporting hyperplane property of convex functions implies that for all $u,v$ \citep{rockafellar1970convex, boyd2004convex},
\begin{equation}
D_{\phi}(u,v)\geq 0.
\end{equation}
Therefore Eq.~\eqref{eq:roleaware} satisfies
\begin{equation}
D_{\theta,P_s,P_t}(x,y)\geq 0
\end{equation}
for all $x,y$. If $\phi$ is strictly convex, then $D_{\phi}(u,v)=0$ if and only if $u=v$. Consequently,
\begin{equation}
D_{\theta,P_s,P_t}(x,y)=0
\quad\Longleftrightarrow\quad
P_sx=P_ty.
\end{equation}
The projected head therefore no longer claims that $D(x,x)=0$ in the original input space. Instead, it has a role-diagonal identity: a pair has zero divergence when the source representation of the first item equals the target representation of the second item. This is appropriate for directed relations, where source and target roles are semantically distinct. \hfill$\square$

\subsection{Quadratic special case and relation to projected metrics}

The proposed head contains familiar distance heads as special cases. This helps clarify which part of the model creates directionality.

\textbf{Proposition 2.}
If the potential is quadratic,
\begin{equation}
\phi(u)=\frac{1}{2}u^{\top}Hu,\quad H\succeq 0,
\end{equation}
then the role-aware Bregman head becomes a role-aware Mahalanobis distance in the projected space:
\begin{equation}
D_{\phi}(P_sx,P_ty)
=
\frac{1}{2}(P_sx-P_ty)^{\top}H(P_sx-P_ty).
\label{eq:quadratic_case}
\end{equation}
If $H=I$, this reduces to a squared Euclidean distance between the source-role representation of $x$ and the target-role representation of $y$.

\textit{Proof.}
For a quadratic potential, $\nabla\phi(v)=Hv$. Substituting into the Bregman definition gives
\begin{align}
D_{\phi}(u,v)
&=\frac{1}{2}u^{\top}Hu-\frac{1}{2}v^{\top}Hv-v^{\top}H(u-v) \nonumber\\
&=\frac{1}{2}(u-v)^{\top}H(u-v).
\end{align}
Taking $u=P_sx$ and $v=P_ty$ gives Eq.~\eqref{eq:quadratic_case}. \hfill$\square$

This special case shows that the proposed model is not merely an unconstrained asymmetric neural scorer. It generalizes a role-aware projected metric by replacing the fixed quadratic potential with a learned convex potential.

\subsection{Source of directional asymmetry}

Although Bregman divergences can be asymmetric even without role projections, the experiments show that this asymmetry does not always align with the annotated source-to-target relation. The role-aware formulation makes the source of directionality explicit. Let
\begin{equation}
a=P_sx,\quad b=P_ty,\quad c=P_sy,\quad d=P_tx.
\end{equation}
The directional gap can be written as
\begin{equation}
G(x,y)=D_{\phi}(a,b)-D_{\phi}(c,d).
\end{equation}

\textbf{Proposition 3.}
Under the quadratic potential in Proposition 2, the directional gap has the exact decomposition
\begin{align}
G(x,y)
&=
\frac{1}{2}(P_sx-P_ty)^{\top}H(P_sx-P_ty) \nonumber\\
&\quad -
\frac{1}{2}(P_sy-P_tx)^{\top}H(P_sy-P_tx).
\label{eq:gap_decomposition}
\end{align}

Equation~\eqref{eq:gap_decomposition} shows that directionality is produced by an interaction between role-specific projections and the geometry induced by $H$. If $P_s=P_t$ and $H$ is symmetric positive semidefinite, the two terms become identical after swapping $x$ and $y$, so the gap vanishes in the quadratic case. Thus, in this interpretable limit, role separation is necessary for directional preference. With a neural convex potential, additional asymmetry can also arise from the non-quadratic shape of $\phi$, but the role maps still determine which representation is evaluated as source and which is evaluated as target.

\subsection{Convexity and local curvature}

For a fixed second argument $v$, $D_{\phi}(u,v)$ is convex in the first argument $u$ because it is the sum of the convex function $\phi(u)$ and a linear term in $u$. For the role-aware head, this means that the divergence is convex in the source-role variable $P_sx$ when the target-role variable $P_ty$ is fixed.

The local geometry is controlled by the Hessian of the potential. If $u=v+\delta$ and $\phi$ is twice differentiable, Taylor expansion around $v$ gives
\begin{equation}
D_{\phi}(v+\delta,v)
\approx
\frac{1}{2}\delta^{\top}\nabla^2\phi(v)\delta.
\end{equation}
For the role-aware head, $\delta=P_sx-P_ty$ describes the local displacement between the source-role and target-role representations. The trace and eigenvalues of $\nabla^2\phi(P_ty)$ therefore quantify local curvature around the target role. In our experiments, the minimum eigenvalues remain positive because the potential includes a strongly convex quadratic component. This supports the interpretation that the learned head does not behave like an unconstrained black-box scorer.

\subsection{Directional gap}

We define the directional gap
\begin{equation}
G(x,y)=D(x,y)-D(y,x).
\end{equation}
For a directed positive pair $(x,y)$, a negative gap means the learned divergence prefers the annotated forward direction. Unlike ranking accuracy, which compares positive and corrupted pairs, the directional gap directly measures whether the head distinguishes the observed relation from its reversal. This quantity is also useful for case studies: pairs with the largest negative gaps are examples where the model expresses strongest directional confidence.

\section{Experiments}

\subsection{Datasets}

We evaluate on five main directed semantic and ontology benchmarks. HyperLex contains graded lexical entailment pairs represented with GloVe embeddings \citep{pennington2014glove}. WordNet contains lexical hierarchy edges also represented with GloVe embeddings. SICK and SNLI contain sentence entailment pairs represented with Sentence-BERT embeddings \citep{reimers2019sentence}. Gene Ontology contains directed biological concept relations represented with Sentence-BERT embeddings of term names and definitions.

Across these datasets, the notation $(x,y)$ always denotes an ordered pair rather than an unordered similarity pair. In HyperLex, $x$ is the more specific lexical item and $y$ is the more general lexical item; the dataset provides a graded lexical entailment score, which we use as relation strength and for filtering/evaluation. In WordNet, $x$ is a child synset and $y$ is a hypernym synset; the label is the existence of a hypernym edge. In SICK and SNLI, $x$ is the premise sentence and $y$ is the hypothesis sentence; entailment pairs are treated as positive directed relations. In Gene Ontology, $x$ is a child GO term and $y$ is a parent GO term connected by an \texttt{is\_a} or \texttt{part\_of} relation. In all cases, the numerical divergence $D(x,y)$ is not observed in the dataset; it is learned from these relation labels by ranking annotated targets above sampled corrupted targets.

We also evaluate OGBL-Citation2 as a larger fixed-feature stress test in which the distance head is applied to fixed node features without a graph neural encoder.

For citation graphs, $x$ is the citing paper and $y$ is the cited paper, and the label is whether the directed citation edge exists. Node features are fixed bag-of-words or provided node attributes. Positive pairs are observed citation edges; negative targets are sampled papers that are not observed as cited by the same source paper.

\subsection{Baselines}

We compare the proposed role-aware Bregman head with symmetric Euclidean, cosine, and Mahalanobis distances; an unstructured MLP scorer; a bilinear asymmetric scorer; a plain ICNN-Bregman head without role projections; order embeddings; and a Poincare-style hyperbolic baseline. The comparison is designed to separate three factors: symmetry versus asymmetry, structured versus unstructured asymmetry, and plain Bregman asymmetry versus role-aware Bregman asymmetry.

\subsection{Evaluation metrics}

Ranking accuracy measures whether a positive pair is scored better than sampled negatives. Direction accuracy measures whether the annotated direction is preferred to the reversed pair. For link-prediction and parent-retrieval evaluation, we also report AUC, average precision, mean reciprocal rank (MRR), and Hits at $K$ (\mbox{Hits@K}), following common ranking-based evaluation practice for link prediction \citep{hu2020open}. MRR is the average reciprocal rank of the true target among candidate targets: if the annotated target is ranked at position $r$, the contribution is $1/r$. Hits at $K$ is the fraction of queries for which the annotated target appears in the top $K$ ranked candidates. Negative divergence rate measures the fraction of evaluated pairs assigned a negative value; for a valid divergence this should be zero. Directional gap statistics and Hessian traces are reported for interpretability.

All main semantic and ontology results use ten random seeds. We report mean and standard deviation. For the key comparison between the proposed role-aware Bregman head and the plain ICNN-Bregman head, we use paired seed-level tests and bootstrap confidence intervals over the seed differences.

\section{Results and Analysis}

\subsection{Main directed semantic and ontology results}

Table~\ref{tab:main} summarizes the central results. The proposed role-aware Bregman head consistently improves direction accuracy over the plain ICNN-Bregman head on all five main datasets. The improvements are large on HyperLex and SNLI, moderate on SICK and WordNet, and smaller but consistent on Gene Ontology. At the same time, the proposed head keeps a zero observed negative divergence rate, unlike the bilinear scorer, which often produces negative values.

\begin{table}[t]
\centering
\small
\caption{Main ten-seed results. Values are mean $\pm$ standard deviation. R-Acc is ranking accuracy; D-Acc is direction accuracy. Bold entries mark the proposed role-aware Bregman head and its key direction/validity measurements, rather than best-in-column values.}
\label{tab:main}
\resizebox{\linewidth}{!}{%
\begin{tabular}{llccc}
\toprule
Dataset & Model & R-Acc & D-Acc & Neg. rate \\
\midrule
\multirow{4}{*}{HyperLex}
& \ICNN & $0.9642\pm0.0094$ & $0.3883\pm0.0394$ & $0.0000$ \\
& \textbf{\PB} & $0.9496\pm0.0184$ & $\mathbf{0.9015\pm0.0204}$ & $\mathbf{0.0000}$ \\
& MLP & $0.9197\pm0.0269$ & $0.9102\pm0.0215$ & $0.0000$ \\
& Order & $0.9672\pm0.0173$ & $0.9124\pm0.0236$ & $0.0000$ \\
\midrule
\multirow{3}{*}{SICK}
& \ICNN & $0.9982\pm0.0024$ & $0.5645\pm0.0190$ & $0.0000$ \\
& \textbf{\PB} & $0.9745\pm0.0064$ & $\mathbf{0.6450\pm0.0161}$ & $\mathbf{0.0000}$ \\
& MLP & $0.9943\pm0.0039$ & $0.6655\pm0.0117$ & $0.0000$ \\
\midrule
\multirow{3}{*}{SNLI}
& \ICNN & $0.9887\pm0.0063$ & $0.6527\pm0.0267$ & $0.0000$ \\
& \textbf{\PB} & $0.9687\pm0.0130$ & $\mathbf{0.8279\pm0.0168}$ & $\mathbf{0.0000}$ \\
& MLP & $0.9806\pm0.0072$ & $0.8730\pm0.0178$ & $0.0000$ \\
\midrule
\multirow{4}{*}{WordNet}
& \ICNN & $0.9061\pm0.0042$ & $0.7906\pm0.0116$ & $0.0000$ \\
& \textbf{\PB} & $0.9087\pm0.0042$ & $\mathbf{0.8380\pm0.0131}$ & $\mathbf{0.0000}$ \\
& MLP & $0.9610\pm0.0025$ & $0.8546\pm0.0111$ & $0.0000$ \\
& Order & $0.9291\pm0.0029$ & $0.8249\pm0.0100$ & $0.0000$ \\
\midrule
\multirow{4}{*}{Gene Ontology}
& \ICNN & $0.9864\pm0.0014$ & $0.9175\pm0.0056$ & $0.0000$ \\
& \textbf{\PB} & $0.9873\pm0.0022$ & $\mathbf{0.9424\pm0.0062}$ & $\mathbf{0.0000}$ \\
& MLP & $0.9949\pm0.0008$ & $0.9444\pm0.0044$ & $0.0000$ \\
& Order & $0.9960\pm0.0006$ & $0.9515\pm0.0053$ & $0.0000$ \\
\bottomrule
\end{tabular}
}
\end{table}

\subsection{Statistical significance}

Table~\ref{tab:sig} reports paired ten-seed comparisons between the role-aware and plain ICNN-Bregman heads. Direction accuracy improves significantly on every main dataset. The same analysis also confirms that the role-aware Bregman head has a significantly lower negative-value rate than the bilinear scorer.

\begin{table}[t]
\centering
\small
\caption{Paired seed-level significance tests for direction accuracy: role-aware Bregman minus plain ICNN-Bregman. CI denotes bootstrap confidence interval over seed differences; $p$-values are from an exact two-sided sign test over the ten paired seeds.}
\label{tab:sig}
\begin{tabular}{lccc}
\toprule
Dataset & Mean difference & 95\% CI & $p$-value \\
\midrule
HyperLex & $+0.5131$ & $[0.4876,0.5394]$ & $0.00195$ \\
SICK & $+0.0804$ & $[0.0680,0.0920]$ & $0.00195$ \\
SNLI & $+0.1752$ & $[0.1606,0.1901]$ & $0.00195$ \\
WordNet & $+0.0474$ & $[0.0390,0.0562]$ & $0.00195$ \\
Gene Ontology & $+0.0249$ & $[0.0219,0.0284]$ & $0.00195$ \\
\bottomrule
\end{tabular}
\vspace{2mm}
\begin{minipage}{0.92\linewidth}
\footnotesize
Note: with ten paired seeds, the exact two-sided sign test is discrete. If all ten seed-level differences favor the same method, the minimum attainable $p$-value is $2/2^{10}=0.001953$, which explains why the reported $p$-values are identical across datasets.
\end{minipage}
\end{table}

\subsection{Projection and direction-weight ablations}

Table~\ref{tab:ablation} reports the ablation evidence. The projection ablation uses HyperLex, SNLI, and WordNet over five seeds and compares plain ICNN-Bregman, shared projection, source-only projection, target-only projection, and the full source-target role-aware head. The results show that role separation is the key design choice. Shared projection variants behave similarly to plain ICNN-Bregman heads and often fail to recover directionality. Source-only and target-only projections improve direction accuracy, showing that the improvement does not come merely from adding parameters. This empirical pattern agrees with the quadratic gap decomposition in Eq.~\eqref{eq:gap_decomposition}: when source and target roles are not separated, the projected quadratic limit cannot express a directional preference between a pair and its reversal. The full source-target variant is a robust default, while target-only projection can be especially strong on hierarchy-like target roles.

The direction-weight sweep in Table~\ref{tab:ablation} shows a clear ranking-direction trade-off. Increasing the weight of the forward-reverse margin improves direction accuracy up to a moderate range. On SNLI and WordNet, very large weights provide little additional direction gain and slightly reduce ranking quality. This supports the interpretation that directionality is not obtained for free; it must be balanced against positive-negative ranking.

\begin{table}[t]
\centering
\small
\caption{Projection and direction-weight ablations over five seeds. The upper panel reports direction accuracy for projection variants. The lower panel reports ranking accuracy / direction accuracy for different direction-loss weights using the full role-aware projected Bregman head.}
\label{tab:ablation}
\resizebox{\linewidth}{!}{%
\begin{tabular}{llccccc}
\toprule
Panel & Dataset & Plain & Shared & Source-only & Target-only & Source-target \\
\midrule
\multirow{3}{*}{Projection D-Acc}
& HyperLex & $0.3080$ & $0.3124$ & $0.7182$ & $\mathbf{0.7562}$ & $0.7299$ \\
& SNLI & $0.6904$ & $0.6952$ & $0.8654$ & $\mathbf{0.8809}$ & $0.8779$ \\
& WordNet & $0.6866$ & $0.6908$ & $0.7306$ & $\mathbf{0.7376}$ & $\mathbf{0.7376}$ \\
\midrule
Panel & Dataset & $\alpha=0$ & $\alpha=0.1$ & $\alpha=0.3$ & $\alpha=1.0$ & $\alpha=3.0$ \\
\midrule
\multirow{3}{*}{R-Acc / D-Acc}
& HyperLex & $0.6715/0.6905$ & $0.6803/0.7007$ & $0.6847/0.7139$ & $0.6934/0.7299$ & $\mathbf{0.6949}/\mathbf{0.7314}$ \\
& SNLI & $0.8955/0.7012$ & $0.9063/0.8301$ & $0.9069/0.8564$ & $\mathbf{0.9104}/\mathbf{0.8779}$ & $0.9003/0.8767$ \\
& WordNet & $0.8304/0.6722$ & $0.8339/0.7164$ & $\mathbf{0.8347}/0.7250$ & $0.8340/\mathbf{0.7376}$ & $0.8311/0.7320$ \\
\bottomrule
\end{tabular}
}
\end{table}

\subsection{Interpretability}

Interpretability is one of the main reasons for using a structured divergence head instead of an unconstrained asymmetric scorer. We therefore analyze the learned head at three levels. The first level is \emph{validity}: because the score is a Bregman divergence in the role-projected space, negative distance values should not occur. The second level is \emph{directional preference}: the directional gap $G(x,y)=D(x,y)-D(y,x)$ indicates whether the model prefers the annotated direction or its reversal. The third level is \emph{local geometry}: the Hessian of the convex potential describes how sensitive the divergence is to perturbations around the target-role representation. These diagnostics operationalize the theoretical analysis in Section~4: the gap measures the observable consequence of the source-target decomposition, while the Hessian summarizes the local convex geometry through which role-projected differences are evaluated.

\begin{figure}[t]
\centering
\includegraphics[width=0.82\linewidth]{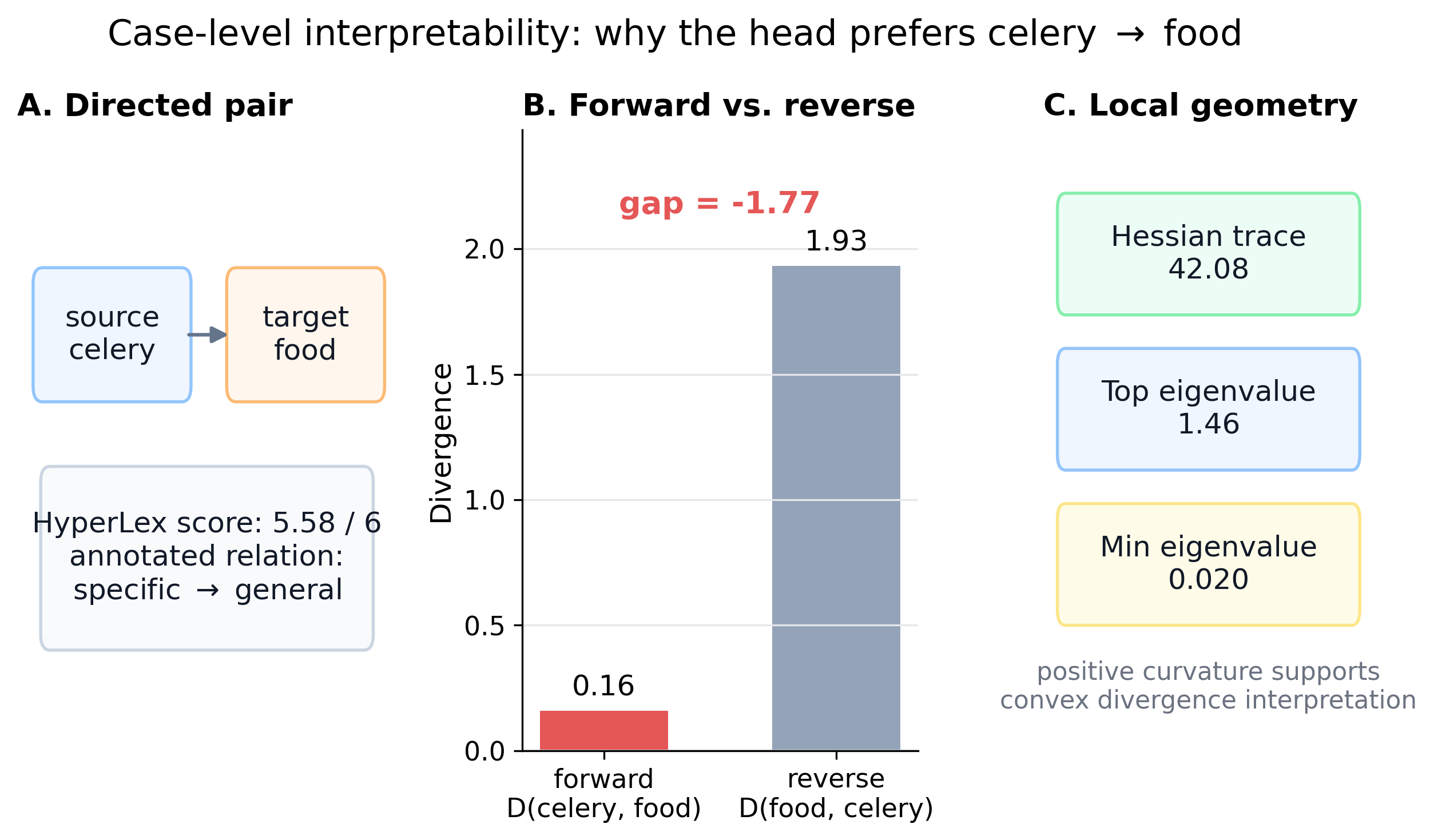}
\caption{Case-level interpretability on HyperLex. For the pair celery $\rightarrow$ food, the role-aware Bregman head assigns a much smaller forward divergence than reverse divergence, producing a negative directional gap. The same example also reports local Hessian diagnostics of the convex potential.}
\label{fig:case_interpretability}
\end{figure}

Table~\ref{tab:interp} and Figure~\ref{fig:case_interpretability} summarize these diagnostics. The directional gap becomes more consistently negative for annotated forward pairs after role-aware projection. On HyperLex, the fraction of pairs with forward-preferred gap increases from $0.3358$ for plain ICNN-Bregman to $0.9124$ for the role-aware head. This is not merely a ranking improvement against sampled negatives; it means that the head learns a directional preference between an observed pair and its reversed counterpart. WordNet and Gene Ontology already have stronger directionality under the plain Bregman head, but role-aware projection still increases the forward-preferred rate and makes the average gap more negative.

The case-level view in Figure~\ref{fig:case_interpretability} shows how this diagnostic can be used in practice. For the HyperLex pair celery $\rightarrow$ food, the learned divergence is $D(\text{celery},\text{food})=0.16$ in the annotated direction and $D(\text{food},\text{celery})=1.93$ in the reverse direction, producing a gap of $-1.77$. This gives an interpretable answer to the question ``why did the model prefer this direction?'': the source-role representation of the specific concept is close to the target-role representation of the general concept, while the reverse role assignment is much farther away. Such examples can be inspected directly, sorted by gap magnitude, or compared with annotation scores.

The Hessian analysis complements the gap analysis. A generic MLP scorer can produce a directional score, but it does not provide a convex potential whose curvature can be inspected. In contrast, the role-aware Bregman head allows local curvature summaries such as the Hessian trace, top eigenvalue, and minimum eigenvalue. The positive minimum eigenvalues observed in the case-level example and aggregate analysis are expected from the strongly convex quadratic component, and they provide a sanity check that the head remains in a convex-divergence regime. The trace values are larger for WordNet and Gene Ontology than for HyperLex, suggesting that the potential uses stronger local curvature on denser or more heterogeneous hierarchy structures. We do not treat the trace as a causal explanation of a prediction, but as a geometric diagnostic of how the learned divergence organizes the projected space.

Together, these diagnostics distinguish the proposed head from both symmetric distances and unconstrained asymmetric scorers. Symmetric distances cannot produce nonzero directional gaps by construction. MLP and bilinear scorers can express directionality, but their scores are not guaranteed to be nonnegative divergences and they do not yield a convex potential for Hessian-based inspection. The role-aware Bregman head therefore provides a middle ground: its predictions are still learned neural scores, but they can be audited through directional gaps, divergence validity, and local convex geometry. In this sense, the interpretability analysis is not an auxiliary visualization step; it is the empirical counterpart of the gap decomposition and curvature characterization.

\begin{table}[t]
\centering
\small
\caption{Directional gap and curvature diagnostics. A higher $G<0$ rate means the model more often prefers the annotated forward direction.}
\label{tab:interp}
\resizebox{\linewidth}{!}{%
\begin{tabular}{llccc}
\toprule
Dataset & Model & Gap mean & $G<0$ rate & Hessian trace \\
\midrule
HyperLex & \ICNN & $0.0001$ & $0.3358$ & -- \\
HyperLex & \PB & $-0.6127$ & $0.9124$ & $42.1213$ \\
WordNet & \ICNN & $-0.1434$ & $0.7810$ & -- \\
WordNet & \PB & $-0.7439$ & $0.8490$ & $94.7032$ \\
Gene Ontology & \ICNN & $-0.7000$ & $0.9160$ & -- \\
Gene Ontology & \PB & $-1.9039$ & $0.9360$ & $83.0153$ \\
\bottomrule
\end{tabular}
}
\end{table}

\subsection{OGB citation stress test}

Table~\ref{tab:ogb} reports OGBL-Citation2 results with fixed node features and no graph neural encoder. Poincare and cosine baselines achieve stronger ranking accuracy and MRR than the proposed head. The role-aware Bregman head slightly improves ranking accuracy and AUC over plain ICNN-Bregman, but does not dominate this large citation setting. The result suggests that large-scale citation prediction requires either stronger input representations or integration with a graph encoder.

\begin{table}[t]
\centering
\small
\caption{OGBL-Citation2 fixed-feature stress test over five seeds.}
\label{tab:ogb}
\resizebox{\linewidth}{!}{%
\begin{tabular}{lcccc}
\toprule
Model & R-Acc & D-Acc & MRR & Neg. rate \\
\midrule
Cosine & $0.9252\pm0.0035$ & $0.5000\pm0.0000$ & $0.6253\pm0.0031$ & $0.0000$ \\
MLP & $0.9205\pm0.0033$ & $0.5464\pm0.0169$ & $0.5691\pm0.0055$ & $0.0000$ \\
\ICNN & $0.8671\pm0.0053$ & $0.5734\pm0.0087$ & $0.5275\pm0.0070$ & $0.0000$ \\
\PB & $0.8730\pm0.0052$ & $0.5532\pm0.0212$ & $0.5233\pm0.0045$ & $0.0000$ \\
Poincare & $0.9341\pm0.0028$ & $0.5000\pm0.0000$ & $0.6299\pm0.0039$ & $0.0000$ \\
Bilinear & $0.8850\pm0.0051$ & $0.5206\pm0.0150$ & $0.4384\pm0.0068$ & $0.7144$ \\
\bottomrule
\end{tabular}
}
\end{table}

\section{Discussion}

The experiments support three conclusions. First, plain Bregman asymmetry is not sufficient in practice. Although Bregman divergences are mathematically asymmetric, the learned head can still fail to align that asymmetry with the annotated relation direction. Role-aware projections address this mismatch by allowing an embedding to take different forms when it acts as a source or target. This conclusion is consistent with the theoretical gap decomposition: the direction score depends not only on the convex potential, but also on how a pair is assigned to source and target roles.

Second, the proposed method should be understood as a structured alternative to generic asymmetric scorers. MLP scorers often achieve strong direction accuracy and can be stronger on pure retrieval metrics. However, they do not provide nonnegative divergence values, projected-space identity, or Hessian-based curvature diagnostics. Conversely, symmetric distances can retrieve semantically or topologically related items well, but they remain direction-blind. The role-aware Bregman head is therefore most attractive when the user wants both directional performance and a distance-like geometric explanation.

Third, the method is not a universal ranking winner. Symmetric distances and hyperbolic baselines remain highly competitive when the input embeddings already encode the task well or when the evaluation mostly rewards undirected neighborhood quality. This is visible in SICK, SNLI, Gene Ontology, and OGB. The contribution is therefore not that role-aware Bregman heads replace all scorers, but that they provide a principled option for directed relations where geometric validity and interpretation matter.

\section{Limitations and Future Work}

The current experiments use fixed embeddings, which isolates the behavior of the comparison head but does not test full end-to-end representation learning. Joint encoder and head training may improve ranking performance, especially on graph benchmarks where the OGB stress test currently uses fixed node features rather than a graph neural encoder. The role projections are mostly linear; nonlinear role maps could increase expressiveness but may weaken interpretability. The empirical study also focuses on head-level comparisons rather than complete task-specific systems. Future work should evaluate the head inside stronger encoders, study robustness to noisy directed labels, and develop generalization guarantees for convex-divergence heads.

\section{Conclusion}

This paper presented a role-aware neural convex divergence head for asymmetric representation learning. By combining source-target role projections with an ICNN-induced Bregman divergence, the method preserves structured nonnegative divergence values while improving directional discrimination over plain ICNN-Bregman heads. The theoretical analysis characterizes how the head retains projected-space Bregman properties and how directional gaps arise from the interaction between role separation and convex geometry. Experiments across lexical, sentence, ontology, and graph benchmarks show that the method is especially useful when asymmetric relations are central and interpretability is desired. The ablation and interpretability analyses further connect the theory to practice: role separation drives directional gains, while gap and Hessian diagnostics make the learned asymmetry inspectable. The results also clarify its limitations: unstructured MLPs and specialized geometric baselines can outperform it on pure ranking metrics. Overall, role-aware neural convex divergence heads provide a practical and interpretable plug-in distance head for directed representation learning.

\section*{CRediT authorship contribution statement}

He Huang: Conceptualization, Methodology, Software, Formal analysis, Investigation, Writing -- original draft, Funding acquisition, Project administration. Lu Shen: Methodology, Investigation, Validation, Resources, Writing -- review and editing, Funding acquisition. Yunfeng Huang: Software, Validation, Formal analysis, Visualization, Writing -- review and editing. Li Qi: Supervision, Funding acquisition, Resources, Writing -- review and editing.

\section*{Declaration of competing interest}

The authors declare that they have no known competing financial interests or personal relationships that could have appeared to influence the work reported in this paper.

\section*{Data availability}

The experiments use publicly available benchmark datasets. The code and scripts for reproducing the reported results are available in the following GitHub repository:
\begin{center}
\footnotesize\href{https://github.com/HeHuangDortmund/Role-Aware-Neural-Convex-Divergence-Heads}{https://github.com/HeHuangDortmund/Role-Aware-Neural-Convex-Divergence-Heads}
\end{center}
Large raw datasets, generated embeddings, and run outputs can be regenerated using the provided scripts.

\section*{Declaration of generative AI and AI-assisted technologies in the manuscript preparation process}

During the preparation of this work, the authors used OpenAI ChatGPT/Codex to assist with language editing, formatting, code review, and manuscript organization. After using these tools, the authors reviewed and edited the content as needed and take full responsibility for the content of the published article.

\section*{Funding}

\begin{sloppypar}
This work was supported by the Science and Technology Research Program of Chongqing Municipal Education Commission [grant numbers KJQN202500828, KJQN202500837] and the Chongqing Technology and Business University High Level Talent Research Project [grant numbers 2556012, 2556008].
\end{sloppypar}

\bibliographystyle{elsarticle-num-names}
\bibliography{references}

@inproceedings{amos2017input,
  title={Input Convex Neural Networks},
  author={Amos, Brandon and Xu, Lei and Kolter, J. Zico},
  booktitle={Proceedings of the International Conference on Machine Learning},
  pages={146--155},
  year={2017}
}

@article{bregman1967relaxation,
  title={The Relaxation Method of Finding the Common Point of Convex Sets and Its Application to the Solution of Problems in Convex Programming},
  author={Bregman, Lev M.},
  journal={USSR Computational Mathematics and Mathematical Physics},
  volume={7},
  number={3},
  pages={200--217},
  year={1967}
}

@article{banerjee2005clustering,
  title={Clustering with Bregman Divergences},
  author={Banerjee, Arindam and Merugu, Srujana and Dhillon, Inderjit S. and Ghosh, Joydeep},
  journal={Journal of Machine Learning Research},
  volume={6},
  pages={1705--1749},
  year={2005}
}

@book{boyd2004convex,
  title={Convex Optimization},
  author={Boyd, Stephen and Vandenberghe, Lieven},
  publisher={Cambridge University Press},
  address={Cambridge},
  year={2004}
}

@inproceedings{bowman2015snli,
  title={A Large Annotated Corpus for Learning Natural Language Inference},
  author={Bowman, Samuel R. and Angeli, Gabor and Potts, Christopher and Manning, Christopher D.},
  booktitle={Proceedings of the Conference on Empirical Methods in Natural Language Processing},
  pages={632--642},
  year={2015}
}

@inproceedings{bordes2013translating,
  title={Translating Embeddings for Modeling Multi-relational Data},
  author={Bordes, Antoine and Usunier, Nicolas and Garcia-Duran, Alberto and Weston, Jason and Yakhnenko, Oksana},
  booktitle={Advances in Neural Information Processing Systems},
  pages={2787--2795},
  year={2013}
}

@inproceedings{cilingir2020deep,
  title={Deep Divergence Learning},
  author={Cilingir, Hatice Kubra and Manzelli, Rachel and Kulis, Brian},
  booktitle={Proceedings of the 37th International Conference on Machine Learning},
  series={Proceedings of Machine Learning Research},
  volume={119},
  pages={2027--2037},
  year={2020}
}

@article{censor1981iterative,
  title={An Iterative Row-Action Method for Interval Convex Programming},
  author={Censor, Yair and Lent, Arnold},
  journal={Journal of Optimization Theory and Applications},
  volume={34},
  number={3},
  pages={321--353},
  year={1981}
}

@inproceedings{davis2007information,
  title={Information-Theoretic Metric Learning},
  author={Davis, Jason V. and Kulis, Brian and Jain, Prateek and Sra, Suvrit and Dhillon, Inderjit S.},
  booktitle={Proceedings of the International Conference on Machine Learning},
  pages={209--216},
  year={2007}
}

@article{geneontology2019resource,
  title={The Gene Ontology Resource: 20 Years and Still GOing Strong},
  author={{The Gene Ontology Consortium}},
  journal={Nucleic Acids Research},
  volume={47},
  number={D1},
  pages={D330--D338},
  year={2019}
}

@inproceedings{hu2020open,
  title={Open Graph Benchmark: Datasets for Machine Learning on Graphs},
  author={Hu, Weihua and Fey, Matthias and Zitnik, Marinka and Dong, Yuxiao and Ren, Hongyu and Liu, Bowen and Catasta, Michele and Leskovec, Jure},
  booktitle={Advances in Neural Information Processing Systems},
  volume={33},
  pages={22118--22133},
  year={2020}
}

@article{kulis2013metric,
  title={Metric Learning: A Survey},
  author={Kulis, Brian},
  journal={Foundations and Trends in Machine Learning},
  volume={5},
  number={4},
  pages={287--364},
  year={2013}
}

@inproceedings{lu2023neural,
  title={Neural Bregman Divergences for Distance Learning},
  author={Lu, Fred and Raff, Edward and Ferraro, Francis},
  booktitle={International Conference on Learning Representations},
  year={2023}
}

@inproceedings{marelli2014sick,
  title={A SICK Cure for the Evaluation of Compositional Distributional Semantic Models},
  author={Marelli, Marco and Menini, Stefano and Baroni, Marco and Bentivogli, Luisa and Bernardi, Raffaella and Zamparelli, Roberto},
  booktitle={Proceedings of the Ninth International Conference on Language Resources and Evaluation},
  pages={216--223},
  year={2014}
}

@article{miller1995wordnet,
  title={WordNet: A Lexical Database for English},
  author={Miller, George A.},
  journal={Communications of the ACM},
  volume={38},
  number={11},
  pages={39--41},
  year={1995}
}

@inproceedings{nickel2017poincare,
  title={Poincare Embeddings for Learning Hierarchical Representations},
  author={Nickel, Maximilian and Kiela, Douwe},
  booktitle={Advances in Neural Information Processing Systems},
  pages={6341--6350},
  year={2017}
}

@inproceedings{pennington2014glove,
  title={GloVe: Global Vectors for Word Representation},
  author={Pennington, Jeffrey and Socher, Richard and Manning, Christopher D.},
  booktitle={Proceedings of the Conference on Empirical Methods in Natural Language Processing},
  pages={1532--1543},
  year={2014}
}

@inproceedings{reimers2019sentence,
  title={Sentence-BERT: Sentence Embeddings Using Siamese BERT-Networks},
  author={Reimers, Nils and Gurevych, Iryna},
  booktitle={Proceedings of the Conference on Empirical Methods in Natural Language Processing},
  pages={3982--3992},
  year={2019}
}

@book{rockafellar1970convex,
  title={Convex Analysis},
  author={Rockafellar, R. Tyrrell},
  publisher={Princeton University Press},
  address={Princeton, NJ},
  year={1970}
}

@inproceedings{siahkamari2020learning,
  title={Learning to Approximate a Bregman Divergence},
  author={Siahkamari, Ali and Xia, Xide and Saligrama, Venkatesh and Castanon, David A. and Kulis, Brian},
  booktitle={Advances in Neural Information Processing Systems},
  volume={33},
  pages={3603--3612},
  year={2020}
}

@inproceedings{sun2019rotate,
  title={RotatE: Knowledge Graph Embedding by Relational Rotation in Complex Space},
  author={Sun, Zhiqing and Deng, Zhi-Hong and Nie, Jian-Yun and Tang, Jian},
  booktitle={International Conference on Learning Representations},
  year={2019}
}

@article{tversky1977features,
  title={Features of Similarity},
  author={Tversky, Amos},
  journal={Psychological Review},
  volume={84},
  number={4},
  pages={327--352},
  year={1977}
}

@inproceedings{vendrov2016order,
  title={Order-Embeddings of Images and Language},
  author={Vendrov, Ivan and Kiros, Ryan and Fidler, Sanja and Urtasun, Raquel},
  booktitle={International Conference on Learning Representations},
  year={2016}
}

@article{vulic2017hyperlex,
  title={HyperLex: A Large-Scale Evaluation of Graded Lexical Entailment},
  author={Vulic, Ivan and Gerz, Daniela and Kiela, Douwe and Hill, Felix and Korhonen, Anna},
  journal={Computational Linguistics},
  volume={43},
  number={4},
  pages={781--835},
  year={2017}
}

@article{weinberger2009distance,
  title={Distance Metric Learning for Large Margin Nearest Neighbor Classification},
  author={Weinberger, Kilian Q. and Saul, Lawrence K.},
  journal={Journal of Machine Learning Research},
  volume={10},
  pages={207--244},
  year={2009}
}

\end{document}